\documentclass[10pt,twocolumn,letterpaper]{article}

\usepackage{cvpr}
\usepackage{times}
\usepackage{epsfig}
\usepackage{graphicx}
\usepackage{amsmath}
\usepackage{amssymb}
\usepackage{multirow}
\usepackage{verbatim}
\usepackage{mathtools}

\DeclareGraphicsExtensions{.pdf,.png,.jpg}

\DeclareMathOperator*{\argmin}{arg\,min}

\newcommand{\Eref}[1]{Eq.~(\ref{#1})}
\newcommand{\Fref}[1]{Fig.~\ref{#1}}
\newcommand{\Sref}[1]{Sec.~\ref{#1}}

\newcommand{\bI}{{I}}
\newcommand{\bS}{{S}}
\newcommand{\bO}{{O}}
\newcommand{\bP}{{P}}
\newcommand{\bC}{{C}}
\newcommand{\bQ}{{Q}}
\newcommand{\bK}{{K}}

\usepackage[pagebackref=false,breaklinks=true,letterpaper=true,colorlinks,bookmarks=false]{hyperref}

\cvprfinalcopy 

\def\cvprPaperID{***} 

\ifcvprfinal\fi

\begin{document}

\title{Automatic Content-Aware Color and Tone Stylization}

\author{Joon-Young Lee\\
	Adobe Research
	\and
	Kalyan Sunkavalli\\
	Adobe Research
	\and
	Zhe Lin\\
	Adobe Research
	\and
	Xiaohui Shen\\
	Adobe Research
	\and
	In So Kweon\\
	KAIST
}

\maketitle

\begin{figure*}
	\includegraphics{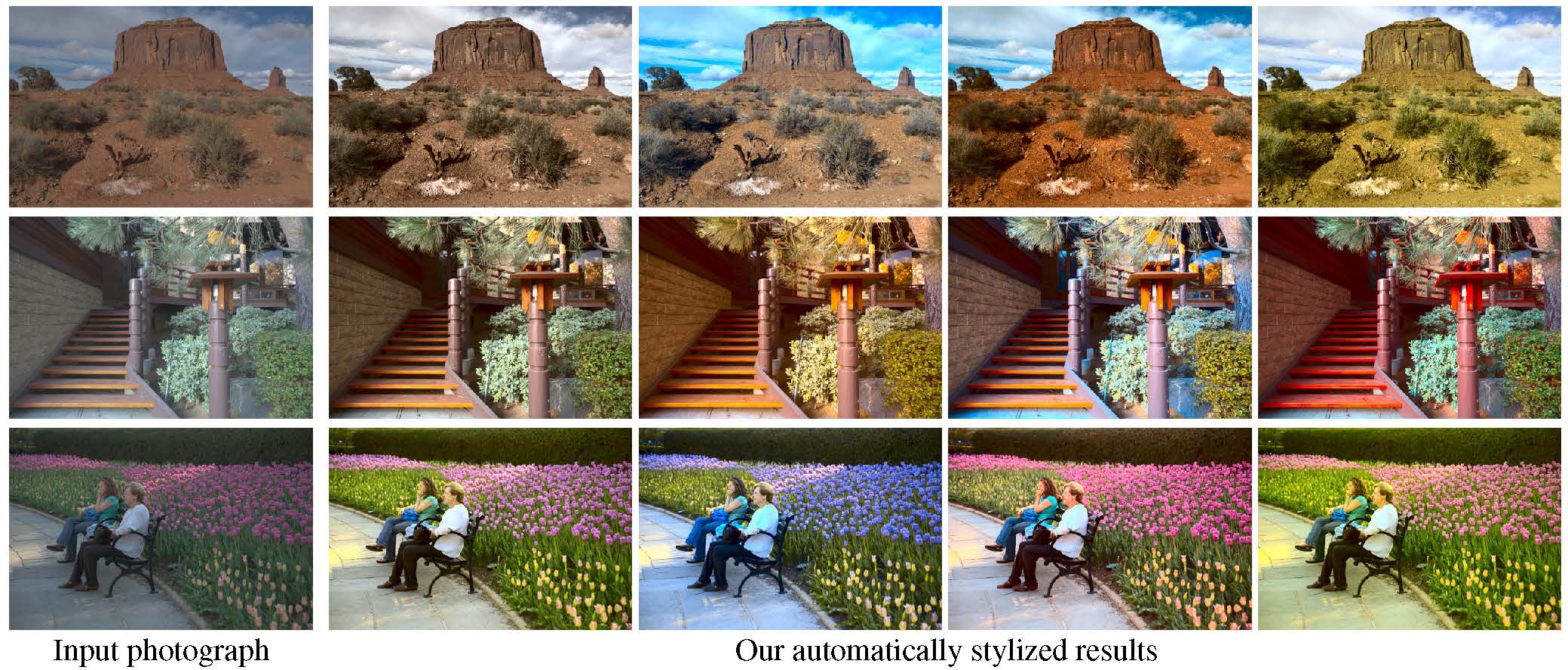}
	
	\caption{Our technique automatically generates a set of different stylistic renditions of an input photograph. We use a combination of semantic and style similarity metrics to learn a style ranking that is specific to the content of the photograph. We sample this ranking to select a subset of diverse styles and robustly transfer their color and tone statistics to the input photograph. This allows us to create stylizations that are diverse, artifact-free, and adapt to content ranging from landscapes to still life to people.}
	\label{fig:teaser}
\end{figure*}

\begin{abstract}
We introduce a new technique that automatically generates diverse, visually compelling stylizations for a photograph in an unsupervised manner. We achieve this by learning style ranking for a given input using a large photo collection and selecting a diverse subset of matching styles for final style transfer. We also propose a novel technique that transfers the global color and tone of the chosen exemplars to the input photograph while avoiding the common visual artifacts produced by the existing style transfer methods. Together, our style selection and transfer techniques produce compelling, artifact-free results on a wide range of input photographs, and a user study shows that our results are preferred over other techniques.
\end{abstract}
\vspace{-3mm}

\section{Introduction}

Photographers often stylize their images by editing their color, contrast and tonal distributions -- a process that requires a significant amount of skill with tools like Adobe Photoshop. Instead, casual users use preset style filters provided by apps like Instagram to stylize their photographs. However, these fixed sets of styles do not work well for every photograph and in many cases, produce poor results.

Example-based style transfer techniques~\cite{reinhard2001color,bae2006two} can transfer the look of a given stylized exemplar to another photograph. However, the quality of these results is tied to the choice of the exemplar used, and the wrong choices often result in visual artifacts. This can be avoided in some cases by directly learning style transforms from input-stylized image pairs~\cite{bychkovsky2011learning,wang2011example,Hwang12Context,Yan14Automatic}. However, these approaches require large amounts of training data, limiting them to a small set of styles. 

Our goal is to make the process of image stylization adaptive by {\em automatically} finding the ``right'' looks  for a photograph (from potentially hundreds or thousands of different styles), and {\em robustly} applying them to produce a diverse set of stylized outputs. In particular, we consider stylizations that can be represented as global transformations of color and luminance. We would also like to do this in an unsupervised manner, without the need for input-stylized example pairs for different content and looks.

We introduce two datasets to derive our stylization technique. The first is our manually curated {\em target style database}, which consists of 1500 stylized exemplar images that capture color and tonal distributions that we consider as good styles. Given an input photograph, we would like to automatically select a subset of these style exemplars that will guarantee good stylization results. We do this by leveraging our second dataset -- a {\em large photo collection} that contains millions of photographs and spans the range of styles and semantic content that we expect in our input photographs (e.g., indoor photographs, urban scenes, landscapes, portraits, etc.). These datasets cannot be used individually for stylization; the style dataset is small and does not span the full content-style space, and the photo collection is not curated and contains both good and poorly-stylized images. The key idea of our work is that we can use the large photo collection to learn a content-to-style mapping and bridge the gap between the source photograph and the target style database. We do this in a completely unsupervised manner, allowing us to easily scale to a large range of image content and photographic styles.

We segment the large photo collection into content-based clusters using semantic features, and learn a ranking of the style exemplars for each cluster by evaluating their style similarities to the images in the cluster. At run time, we determine the semantic clusters nearest to the input photograph, retrieve their corresponding stylized exemplar rankings, and sample this set to obtain a diverse subset of relevant style exemplars. 

We propose a new robust technique to transfer the global color and tone statistics of the chosen exemplars to the input photo. Doing this using previous techniques can produce artifacts, especially when the exemplar and input statistics are very disparate. We use regularized color and tone mapping functions, and use a face-specific luminance correction step to minimize artifacts in the final results. \Fref{fig:teaser} show our stylization results on three example images.

We introduce a new benchmark dataset of 55 images with manually stylized results created by an artist. We compare our style selection method with other variants as well as the artist's results through a blind user study. We also evaluate the performance of a number of current statistics-based style transfer techniques on this dataset, and show that our style transfer technique produces better results than all of them. To the best of our knowledge, this is the first extensive quantitative evaluation of these methods. 

The technical contributions of our work include: 
\begin{enumerate}
\item A robust style transfer method that captures a wide range of looks while avoiding image artifacts,
\item An unsupervised method to learn a content-specific style ranking using semantic and style similarity,
\item A style selection method to sample the ranked styles to ensure both diversity and quality in the results, and
\item A new benchmark dataset with professional stylizations and a comprehensive user evaluation of various style selection and transfer techniques.
\end{enumerate}

\begin{figure*}[t]
	\setlength{\tabcolsep}{1pt}
	\centering
	\includegraphics{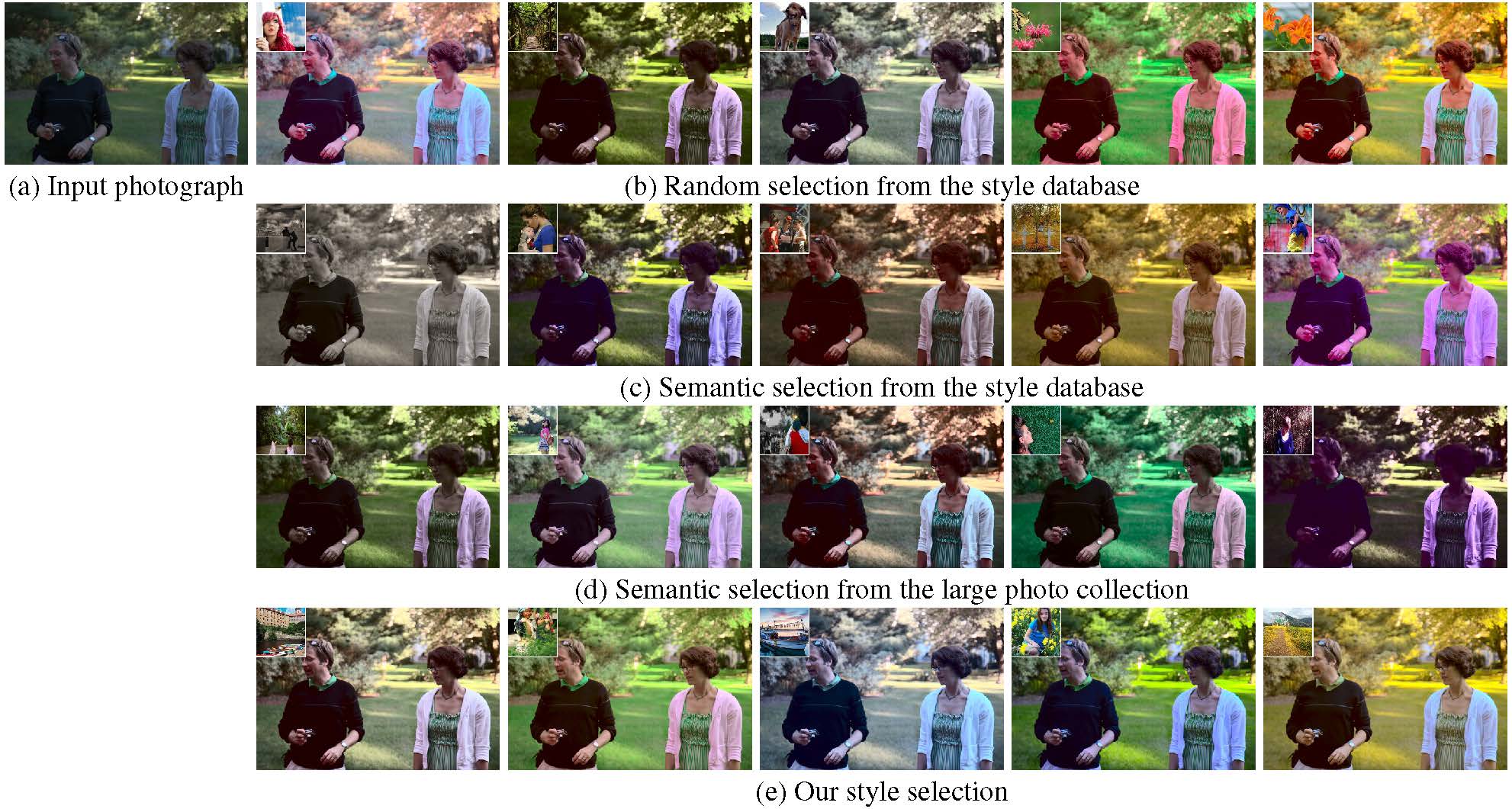}
	\caption{Stylization results with different choices of the exemplar images. All exemplars are shown in insets in the top-left corner.}
	\label{fig:search_comparisons}
\end{figure*}

\section{Related Work}

\paragraph{Example-based Style Transfer} 
One popular approach for image stylization is to transfer the style of an exemplar image to the input image. 
This approach was pioneered by Reinhard et al.~\cite{reinhard2001color} who transferred color between images by matching the statistics of their color distributions. 
There are several subsequent work~\cite{tai2007soft, pitie2005n, pitie2007linear, pouli2011progressive} that improves this technique.
All these techniques are designed to match the input and exemplar color distributions robust to outliers. 
Subsequent work has improved on this technique by using soft-segmentation~\cite{tai2007soft}, multi-dimensional histogram matching~\cite{pitie2005n}, minimal displacement mapping~\cite{pitie2007linear}, and histogram reshaping~\cite{pouli2011progressive}. 
All these techniques are designed to match the input and exemplar color distributions while remaining robust to outliers. 
Instead of transferring color distributions, correspondence-based methods compute (potentially non-linear) color transfer functions from pixel correspondences between the input and exemplar images that are either automatically estimated~\cite{hacohen2011non,hwang:cvpr14} or specified by the user~\cite{an2010user}. 
Example-based color transfer techniques have also been used for video grading~\cite{Bonneel13Video}, realistic compositing~\cite{Johnson11CG2Real,Xue12Realism}, and transferring attributes like time of day to photographs~\cite{Shih13:Time,Laffont14:Attributes}.
Please refer to \cite{xu2010performance, faridul2014survey} for a detailed survey of different color transfer methods. 
We base our chrominance transfer function on the work of Piti{\'e} et al.~\cite{pitie2007linear} but add a regularization term to make it robust to large differences in the color distributions being matched. 

Style transfer techniques also match the contrast and tone between images. 
This is done by manipulating the luminance of the photograph using histogram matching, or applying a parametric tone-mapping curve like a gamma curve or an S-curve~\cite{kang2010personalization}. 
Bae et al.~\cite{bae2006two} propose a two-scale technique to transfer both global and local contrast. 
Aubry et al.~\cite{Aubry14LocalLaplacian} demonstrate the use of local Laplacian pyramids for contrast and tone transfer. 
Shih et al.~\cite{Shih14Portrait} use a multi-scale local contrast transfer technique to stylize portrait photographs. 
We propose a parametric luminance reshaping curve that is designed to be smooth and avoids artifacts in the results. In addition, we propose a face luminance correction method that is specifically designed to avoid artifacts for portrait shots.

\paragraph{Learning-based Stylization and Enhancement} 
Another approach for image stylization is to use supervised methods to learn style mapping functions from data consisting of input-stylized image pairs. 
Wang et al.~\cite{wang2011example} introduce a method to learn piece-wise smooth non-linear color mappings from image pairs.
Yan et al.~\cite{Yan14Automatic} uses deep neural networks to learn local nonlinear transfer functions for a variety of photographic effects.
There are also several automatic learning-based enhancement techniques.
Kang et al.~\cite{kang2010personalization} present a personalized image enhancement framework using distance metric learning.
It was extended by \cite{caicedo2011collaborative}, which proposes collaborative personalization.
Bychkovsky et al.~\cite{bychkovsky2011learning} build a reference dataset of input-output image pairs.
Hwang et al.~\cite{Hwang12Context} propose a context-based local image enhancement method.
Yan et al.~\cite{yan2014learning} account for the intermediate decisions of a user in the editing process.
While these learning-based methods show impressive adjustment results, collecting training data and generalizing them to a large number of styles is very challenging.
In contrast, our technique to learn content-specific style rankings is completely unsupervised and easily generalizes to a large number of content and style classes.

Our technique is similar in spirit to two papers that leverage large image collections to restore/stylize the color and tone of photographs. 
Dale et al.~\cite{dale2009restore} find visually similar images in a large photo collection, and use their aggregate color and tone statistics to restore the input photograph. This aggregation causes a regression to the mean that is appropriate for image restoration but not stylization.
Liu et al.~\cite{liu2014autostyle} use a user-specified keyword to search for images that are used to stylize the input photo. The final results are highly dependent on the choice of the keyword and it can be challenging to predict the right keywords to stylize a photograph. Our technique automatically predicts the right styles for the input photograph.

\begin{figure*}[t]
	\centering
	\includegraphics{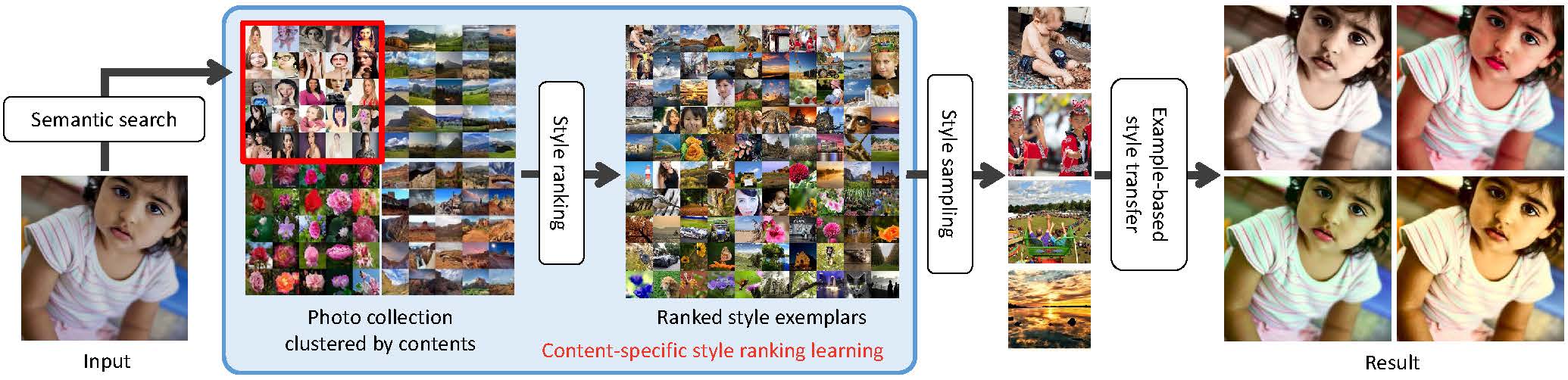}
	\caption{The overall framework of our system.}
	\label{fig:framework}
\end{figure*}

\section{Overview}
\label{sec:overview}

Given an input photograph, $\bI$, our goal is to automatically create a set of $k$ stylized outputs $\bO_1, \bO_2, \cdots, \bO_k$. In particular, we focus on stylizations that can be represented as global transformations of the input color and luminance values. The styles we are interested in are captured by a curated set of exemplar images $\bS_1, \bS_2, \cdots, \bS_n, (n >> k)$. Using images as style examples makes it intuitive for users to specify the looks they are interested in.

We use an example-based style transfer algorithm to transfer the look of a given exemplar image to the input photograph. While example-based techniques can produce compelling results~\cite{faridul2014survey}, they often cause visual artifacts when there are strong differences in the input and exemplar images being processed. In this work, we develop regularized global color and tone mapping functions (\Sref{sec:style_transfer}) that are expressive enough to capture a wide range of effects, but sufficiently constrained to avoid such artifacts.

The quality of the stylized result $\bO$ is also closely tied to the choice of the exemplar $\bS$. Using an outdoor landscape image, for example, to stylize a portrait could lead to poor transfer results (see \Fref{fig:search_comparisons}(b)). It is therefore important to choose the ``right'' set of exemplar images based on the content of the input photograph. We use a semantic similarity metric -- that we learned using a convolutional neural network (CNN) -- to match images with similar content. Given this semantic similarity measure, one approach would be to use it directly to find exemplar images with content similar to an input photograph and stylize it. However, the curated exemplar dataset is limited and unlikely to contain style examples for every content class. Using the semantic similarity metric to find the closest stylized exemplar to an input photograph will not guarantee a good match, and as illustrated in \Fref{fig:search_comparisons}(c), could lead to poor stylizations.

In order to learn a content-specific style ranking, we crawl a large collection of Flickr interesting photos $\bP_1, \bP_2, \cdots, \bP_m, (m >> n)$  that cover a wide range of different content with varying styles and levels of quality. A straightforward way of stylizing an input photograph could be to use the semantic similarity measure to directly find matching images from this large collection and transfer their statistics to the input photograph. However, this large collection of photos is not manually curated, and contains images of both good and bad quality. Performing style transfer using the low-quality photographs in the database can lead to poor stylizations, as shown in \Fref{fig:search_comparisons}(d). While these results can be improved by curating the photo collection, this is an infeasible task given the size of the database. 

We leverage the large photo collection to learn a style ranking for each content class in an unsupervised way. We cluster the photo collection into a set of semantic classes using the semantic similarity metric (\Sref{sec:semantic_clustering}). For each image in a semantic class, we vote for the best matching stylized exemplar using a style similarity metric (\Sref{sec:style_ranking}). We aggregate these votes across all the images in the class to build a content-specific ranking of the stylized exemplars. 

At run time, we match an input photograph to its closest semantic classes and use the pre-computed style ranking for these classes to choose the exemplars. We use a greedy sampling technique to ensure a diverse set of examples (\Sref{sec:style_sampling}), and transfer the statistics of the sampled exemplars to the input photograph using our robust example-based transfer technique. As shown in \Fref{fig:search_comparisons}(e), our style selection technique chooses stylized exemplars that are not necessarily semantically similar to the input photograph, yet have the ``right'' color and tone statistics to transfer, and produces results that are significantly better than the approaches of directly searching for semantically similar images in the style database or the photo collection. \Fref{fig:framework} illustrates the overall framework of our stylization system.

\begin{figure*}
	\setlength{\tabcolsep}{1pt}
	\centering
	\includegraphics{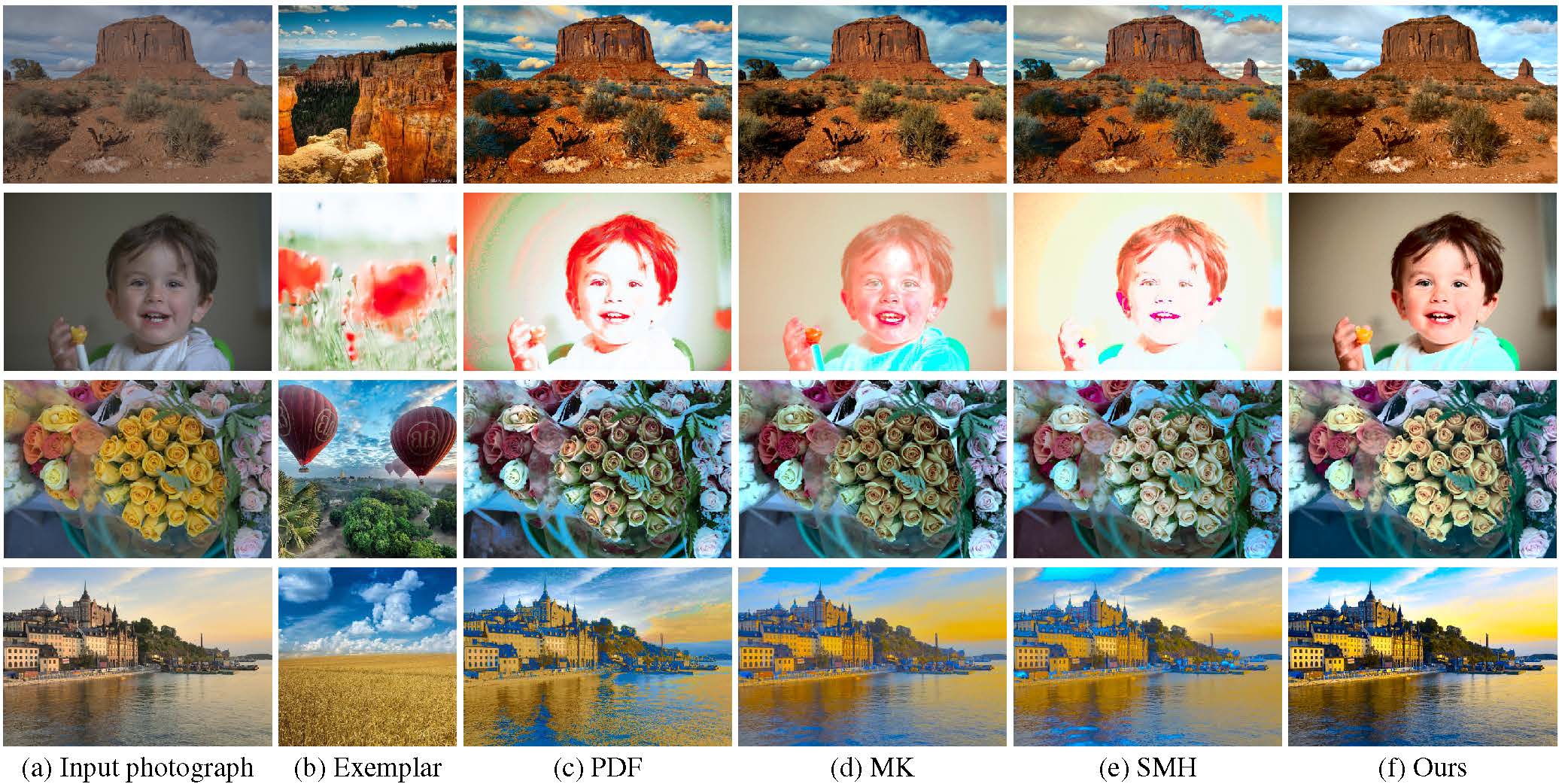}
	\caption{Examples of our style transfer results compared with previous statistics-based transfer methods. Exemplars are shown in insets in the top-left corner of input images.}
	\label{fig:compare_style_transfer}
\end{figure*}

\section{Robust Example-based Style Transfer}
\label{sec:style_transfer}

We stylize an input photograph, $I$, by applying global transforms to match its color and tonal statistics to those of a style example, $S$. 
This space of transformations encompasses a wide range of stylizations that artists use, including color mixing, hue and saturation shifts, and non-linear tone adjustments. 
While a very flexible transfer model can capture a wide range of photographic looks, it is also important that it can be robustly estimated and does not cause artifacts; this is particularly important in our case, where the images being mapped may differ significantly in their content. 
With this in mind, we design color and contrast mapping functions that are regularized to avoid artifacts. 

To effectively stylize images with global transforms, we first compress the dynamic ranges of the two images using a $\gamma$ ($=2.2$) mapping and convert the images into the CIELab colorspace (because it decorrelates the different channels well). Then, we stretch the luminance (L channel) to cover the full dynamic range after clipping both the minimum and the maximum 0.5 percent pixels of luminance levels, and apply different transfer functions to the luminance and chrominance components.

\paragraph{Chrominance}
Our color transfer method maps the statistics of the chrominance channels of the two images. 
We model the chrominance distribution of an image using a multivariate Gaussian, and find a transfer function that creates the output image $\bO$ by mapping the Gaussian statistics $\mathcal{N}_\bS(\mu_\bS, \Sigma_\bS)$ of the style exemplar $\bS$ to the Gaussian statistics $\mathcal{N}_\bI(\mu_\bI, \Sigma_\bI)$ of the input image $\bI$ as:
\begin{equation}
c_\bO(x) = T(c_I(x)-\mu_\bI) + \mu_\bS \quad \text{s.t.} \quad T \Sigma_\bI T^\top = \Sigma_\bS,
\label{eq:gaussian_transfer}
\end{equation}
where $T$ is a linear transformation that maps chrominance between the images and $c(x)$ is the chrominance at pixel $x$. Following Piti{\'e} et al.~\cite{pitie2007linear}, we solve for the color transform using the following closed form solution:
\begin{equation}
T=\Sigma_{\bI}^{-1/2} \left(\Sigma_{\bI}^{1/2}\Sigma_\bS\Sigma_{\bI}^{1/2}\right)^{1/2} \Sigma_{\bI}^{-1/2}.
\label{eq:pitie_transfer}
\end{equation}
This solution is unstable for low input covariance values, leading to color artifacts when the input has low color variation. To avoid this, we regularize this solution by clipping diagonal elements of $\Sigma_{\bI}$ as:
\begin{equation}
\Sigma_{\bI}' = \max(\Sigma_\bI, \lambda_r \mathbb{I}),
\end{equation}
and substitute it into \Eref{eq:pitie_transfer}. Here $\mathbb{I}$ is an identity matrix. This formulation has the advantage that it only regularizes colors channels with low variation without affecting the others. We use a regularization of $\lambda_r=7.5$.

\paragraph{Luminance}
We match contrast and tone using histogram matching between the luminance channels of the input and style exemplar images. Direct histogram matching typically results in arbitrary transfer functions and may produce artifacts due to non-smooth mapping or excessive stretching/compressing of the luminance values. Instead, we design a new parametric model of luminance mapping that allows for strong expressiveness and regularization simultaneously. Our transfer function is defined as:
\begin{equation}
l_\bO(x) = g(l_\bI(x)) = \frac{\arctan(\frac{m}{\delta}) + \arctan(\frac{l_\bI(x)-m}{\delta}) }{ \arctan(\frac{m}{\delta}) + \arctan(\frac{1-m}{\delta}) },
\label{eq:lumi_mapping}
\end{equation}
where $l_\bI(x)$ and $l_\bO(x)$ are the input and output luminance respectively, and $m$ and $\delta$ are the two parameters of the mapping function. $m$ determines the inflection point of the mapping function and $\delta$ determines the degree of luminance stretching around the inflection point. This parametric function can represent a diverse set of tone mapping curves and we can easily control the degree of stretching/compressing of tone. Since the derivative of \Eref{eq:lumi_mapping} is always positive and continuous, it is guaranteed to be a smooth and monotonically increasing curve. This ensures that this mapping function generates a proper luminance mapping curve for any set of parameters.

We extract a luminance feature, $L$, that represents the luminance histogram with uniformly sampled percentiles of the luminance cumulative distribution function (we use 32 samples). We estimate the tone-mapping parameters by minimizing the cost function:
\begin{multline}
(\hat{m},\hat{\delta}) = \argmin_{m,\delta} \lVert g(L_{\bI}) - \tilde{L} \rVert^{2},\\
\text{s.t.} \quad \tilde{L} = {L}_{\bI} + \left( {L}_{\bS} - {L}_{\bI} \right) \frac{\tau}{\min(\tau, |{L}_{\bS} - {L}_{\bI}|_\infty)},
\label{eq:est_tone_param}
\end{multline}
where ${L}_{\bI}$ and ${L}_{\bS}$ represent the the input and style luminance features, respectively. 
$\tilde{L}$ is an interpolation of the input and exemplar luminance features and represents how closely we want to match the exemplar luminance distribution. We set $\tau$ to $0.4$ and minimize this cost using parameter sweeping in a branch-and-bound scheme.

\Fref{fig:compare_style_transfer} compares the quality of our style transfer method against three recent methods: the N-dimensional histogram matching technique of Piti{\'e} et al.~\cite{pitie2005n}, the linear Monge-Kantarovich solution of Piti{\'e} and Kokaram~\cite{pitie2007linear}, and the three-band method of Bonneel et al.~\cite{Bonneel13Video}. While each of these algorithms has its strengths, only our method consistently produces visually compelling results without any artifacts. We further evaluate all these methods via a comprehensive user study in \Sref{sec:results}.

\begin{figure}
	\setlength{\tabcolsep}{1pt}
	\centering
	\includegraphics{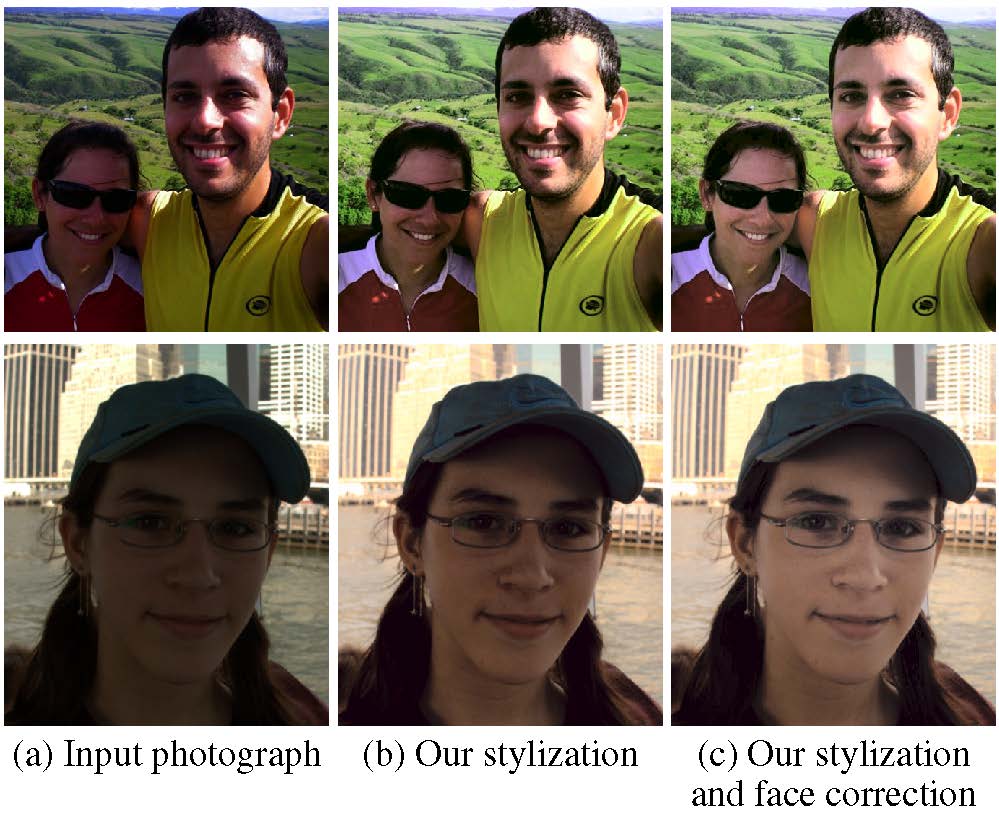}
	\caption{Face exposure correction.}
	\label{fig:face_exposure_correction}
\end{figure}

\paragraph{Face exposure correction}
In the process of transferring tonal distributions, our luminance mapping method can over-darken some regions. When this happens to faces, it detracts from the quality of the result, as humans are sensitive to facial appearance. We fix this using a face-specific luminance correction. We detect face regions in the input image, given by center $p$ and radius $r$, using the OpenCV face detector. If the median luminance in a face region, $\bar{l}$, is lower than a threshold $l_{th}$, we correct the luminance as:
\begin{align}
\hat{l}(x) \;=\;& (1-w(x))*l(x) + w*l(x)^\gamma \;\; \text{if} \; \bar{l} < l_{\,\textrm{th}}, \nonumber \\
      w(x) \;=\;& \exp(-\alpha_r \lVert (x-p)/r \rVert^2) \exp(-\alpha_c \lVert c-\bar{c} \rVert^2), \nonumber \\
    \gamma \;=\;& \max(\gamma_{\,\textrm{th}}, 0.65*\bar{l}/l_{\,\textrm{th}}).
\end{align}
This technique applies a simple $\gamma$-correction to the luminance, where $\gamma_{\,\textrm{th}}$ determines the maximum level of exposure correction. We would like to apply it to the entire face; however, the face region is given by a coarse box and applying the correction to the entire box will produce artifacts. Instead we interpolate the corrected luminance with the original luminance using weights $w(x)$. We compute these weights based on spatial distance from the face center, and chrominance distance from the median face chrominance value, $\bar{c}$ (to capture the color of the skin). $\alpha_r$ and $\alpha_c$ are normalization parameters that control the weights of the spatial and chrominance kernels respectively. We set $\{\gamma_{\,\textrm{th}}, \alpha_r, \alpha_c\}$ to $\{0.5, 0.45, 0.001\}$. \Fref{fig:face_exposure_correction} shows an example of our face exposure correction results.

\section{Content-aware Style Selection}
\label{sec:style_selection}

Given the target style database\footnote{a curated dataset of 1500 exemplar style images}, we can use the method described in \Sref{sec:style_transfer} to transfer the photographic style of a style exemplar to an input photograph. 
However, as noted in \Sref{sec:overview} and illustrated in \Fref{fig:search_comparisons}, it is important that we choose the right set of style exemplars. 
Motivated by the fact that images with different semantic content require different styles, we attempt to learn the set of good styles (or their ranking) for each type of semantic content separately. 

To achieve this, we prepare a large photo collection consisting of one million photographs downloaded from Flickr's daily interesting photograph collection\footnote{https://www.flickr.com/services/api/flickr.interestingness.getList.html}.  
As noted in \Sref{sec:overview}, the curated style dataset does not contain examples for all content classes and cannot be directly used to stylize a photograph. 
However, by leveraging the large photo collection, we can learn style rankings of the curated style dataset even for content classes that are not represented in it. 

The large photo collection captures a joint distribution of content and styles. 
We use a semantic descriptor (\Sref{sec:semantic_clustering}) to cluster the training collection into content classes. 
The semantic feature has a degree of invariance to style, and as a result each class contains images of very similar content but with a variety of different styles, both good and bad. 
This distribution of styles within each content class allows us to learn how compatible a style is with a content class.
The style-to-content compatibility is specifically learned via a simple style-based voting scheme (\Sref{sec:style_ranking}) that evaluates how similar each style exemplar is to the images in the content cluster; style exemplars that occur often are deemed to be better suited to that content class, and conversely, those that occur infrequently are not considered compatible. 

In the on-line phase, we determine the content class of an input photograph and retrieve its pre-computed style ranking. 
We sample this style ranking (\Sref{sec:style_sampling}) to obtain a small set of diverse style images and compute the final results using our style transfer technique~(\Sref{sec:style_transfer}).

\begin{figure}
	\setlength{\tabcolsep}{1pt}
	\centering
	\includegraphics[width=\linewidth]{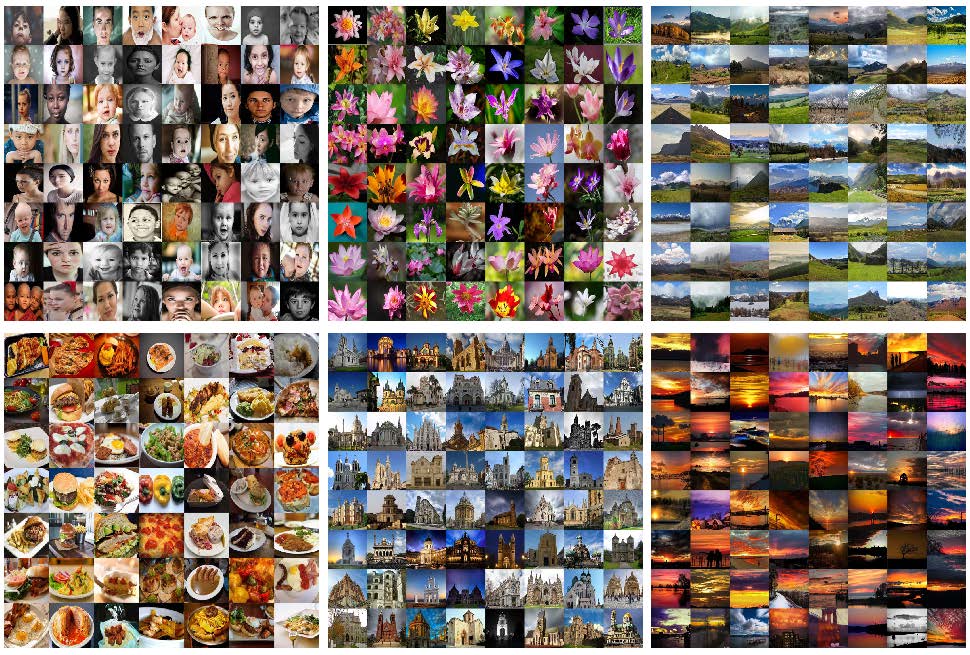}
	\caption{Examples of semantic clusters.}
	\label{fig:semantic_clusters}
\end{figure}

\subsection{Semantic clustering}
\label{sec:semantic_clustering}

Inspired by recent breakthroughs in the use of CNN~\cite{Krizhevsky12CNN}, we represent the semantic information of an image using a CNN feature, trained on the ImageNet dataset~\cite{deng2009imagenet}. We modified the {\em CaffeNet}~\cite{Jia2013caffe} to have fewer nodes in the fully-connected layers and fine-tuned the modified network. This results in a $512$-dimensional feature vector for each image. We empirically found that this smaller CNN captures more style diversity in each content cluster compared to the original {\em CaffeNet} or {\em AlexNet}~\cite{deng2009imagenet} which sometimes ``oversegments'' content into clusters with low style variation.

We perform $k$-means clustering on the CNN feature vectors for each image in the large photo collection to obtain semantic content clusters. A small number of clusters leads to different content classes being grouped in the same cluster, while a large number of clusters lead to the style variations of the same content class of images being split into different clusters. In our experiments, we found that using $1000$ clusters was a good balance between these two aspects.

\Fref{fig:semantic_clusters} shows images from six different semantic clusters. The images in a single cluster share semantically similar content but have diverse appearances (including both good and bad styles). These intra-class style variations allow us to learn the space of relevant styles for each class.

\begin{figure*}[t]
	\setlength{\tabcolsep}{1pt}
	\centering
	\includegraphics[width=\linewidth]{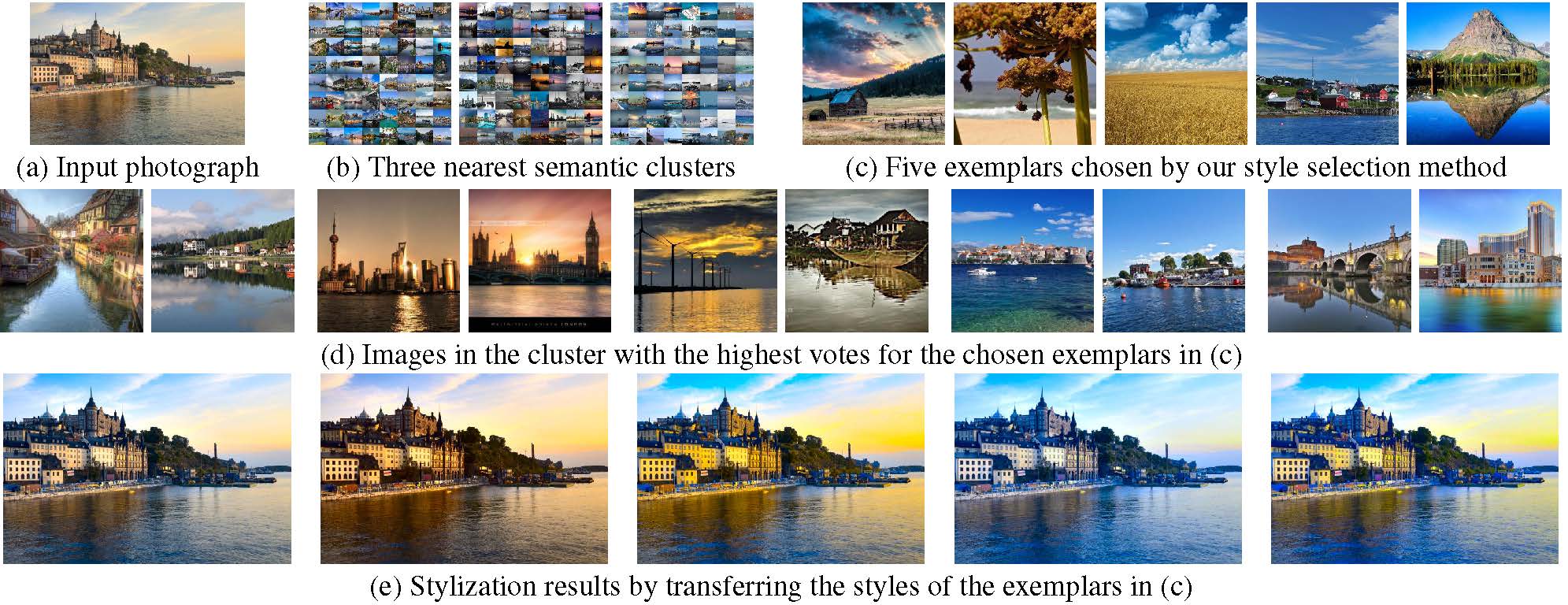}
	\caption{Intermediate steps of style selection. The input (a) can be semantically different from the selected exemplars (c) (second and third example especially). However, the cluster images with the highest votes for these style exemplars (d), are both semantically similar to the input and stylistically similar to the chosen exemplars. This ensures input-exemplar compatibility and leads to artifact-free stylizations (e).}
	\label{fig:result_pipeline}
\end{figure*}

\subsection{Style ranking}
\label{sec:style_ranking}

To choose the best style example for each semantic cluster, we compute style similarity between each style example and the images in a cluster, and use this measure to rank the styles for that cluster. As explained in \Sref{sec:style_transfer}, we represent a photograph's style using chrominance and luminance statistics. Following this, we define the style similarity measure between cluster photograph $\bP$ and style image $\bS$ as:
\begin{equation}
\mathcal{R}(\bP, \bS) =  \exp\left( -\frac{ \mathcal{D}_{e}({L}_\bP, {L}_\bS)^2 }{\lambda_l} \right) \exp\left( -\frac{\mathcal{D}_{h}(\mathcal{N}_\bP, \mathcal{N}_\bS)^2}{\lambda_c} \right),
\end{equation}
where $\mathcal{D}_e$ represents the Euclidean distance between the two luminance features, and $\lambda_l$ and $\lambda_c$ are normalization parameters. We set $\lambda_l=0.005$ and $\lambda_c=0.05$ to generate all our results.
$\mathcal{D}_h$ is the Hellinger distance~\cite{pollard2002user} defined as:
\begin{multline}
\mathcal{D}_h\left(\mathcal{N}_\bP, \mathcal{N}_\bS\right) = 1 - \frac{ |\Sigma_{\bP} \Sigma_{\bS}|^{1/4} }{ |\bar\Sigma|^{1/2} }\exp \left( -\frac{1}{8}\bar\mu^{\top} \bar\Sigma^{-1} \bar\mu \right) \\
\text{s.t} \quad \bar\mu=|\mu_{\bP}-\mu_{\bS}| + \epsilon, \quad \bar\Sigma \;=\; \frac{ \Sigma_{\bP} + \Sigma_{\bS}}{2},
\end{multline}
where $\mathcal{N}_{\bP} = (\mu_\bP, \Sigma_\bP)$ are the multivariate Gaussian statistics of chrominance channel for an image. 
We chose the Hellinger distance to measure the overlap between two distributions because it strongly penalizes large differences in covariance even if the means are close enough. 
$\epsilon=1$ is added to the difference between the means to additionally penalize small covariance images.

We measure the compatibility of a stylized exemplar $\bS$, with a semantic cluster $\bC_\bK$, by aggregating the style similarity measure over all the images in the cluster as
\begin{equation}
\bar{\mathcal{R}}_\bK(\bS) = \sum_{\bP \in \bC_\bK}{ \mathcal{R}(\bP, \bS) }.
\label{eqn:aggregate}
\end{equation}
For each semantic cluster, we compute $\bar{\mathcal{R}}$ for all the style exemplars and determine the style example ranking by sorting $\bar{\mathcal{R}}$ in decreasing order.
This voting scheme measures how often a particular exemplar's color and tonal statistics occurs in the semantic cluster.
Poorly stylized cluster images are implicitly filtered out because they do not vote for any style exemplar. 
Meanwhile, well stylized images in the cluster vote for their corresponding exemplars, giving us a ``histogram'' of the style exemplars for that cluster. 

Figs.~\ref{fig:framework} and~\ref{fig:result_pipeline} show the results of each stage of our stylization pipeline. As these figures illustrate, our semantic similarity term is able to find clusters with semantically similar content (see \Fref{fig:result_pipeline}(b)). Our technique does not require the selected style exemplars to be semantically similar to the input image (see \Fref{fig:result_pipeline}(c)). While this might seem counter-intuitive, the final stylized results do not suffer from any artifacts because the highly-ranked styles have the same style characteristics as a large number of ``auxiliary exemplars'' in the training photo collection that, in turn, share the same content as the input (see \Fref{fig:result_pipeline}(d)). This is an important property of our style selection scheme, and is what allows it to generalize a small style dataset to arbitrary content. 

We also experimented with an alternative way of ranking styles based on a weighted combination of style and semantic similarity between the curated dataset and the large photo collection. However, our empirical experiments showed that it was consistently worse than relying solely on the style similarity due to the lack of semantically similar examples with diverse styles in the curated dataset. We evaluate our style selection criteria against other candidate methods via a user study in \Sref{sec:results}.

\begin{figure*}[t]
\setlength{\tabcolsep}{1pt}
\centering
\includegraphics{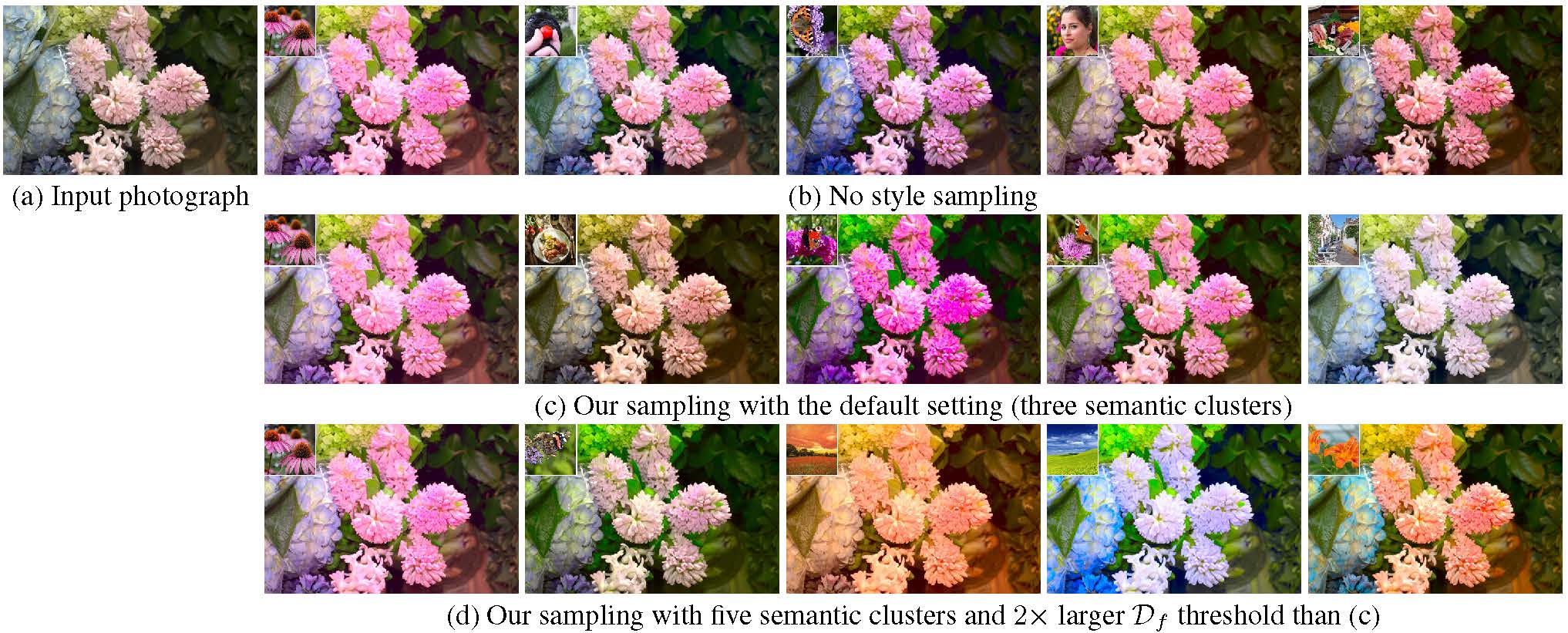}
\caption{Results according to different style sampling strategies (example images in insets). Directly using the top-ranked style examples from the learned ranking can lead to similar results (b). Our sampling strategy combines styles from multiple semantic clusters and enforces a certain style diversity threshold (c). Increasing the number of clusters and the threshold increases diversity (d).}
\label{fig:compare_sampling}
\vspace{-3mm}
\end{figure*}

\subsection{Style sampling}
\label{sec:style_sampling}

Given an input photograph, we can extract its semantic feature and assign it to the nearest semantic cluster.
We can retrieve the pre-computed style ranking for this cluster and use the top $k$ style images to create a set of $k$ stylized renditions of the input photograph. However, this strategy could lead to outputs that are similar to each other. In order to improve the diversity of styles in the final results, we propose the following multi-cluster style sampling scheme. 

Adjacent semantic clusters usually share similar high-level semantics but different low-level features such as object scale, color, and tone. Therefore we propose using {\em multiple} nearest semantic clusters to capture more diversity. We merge the style lists for the chosen semantic clusters and order them by the aggregate similarity measure (\Eref{eqn:aggregate}). To avoid redundant styles, we sample this merged style list in order (starting with the top-ranked one) and discard styles that are within a specified threshold distance from the styles that have already been chosen.

We define a new similarity measure for this sampling process that computes the squared Fr\'echet distance~\cite{dowson1982frechet}:
\begin{multline}
\mathcal{D}_f(\mathcal{N}_\bP, \mathcal{N}_\bQ)  =  \\
\sqrt{\lVert \mu_\bP - \mu_\bQ \rVert^2 + \text{tr}[\Sigma_\bP + \Sigma_\bQ  - 2(\Sigma_\bP \Sigma_\bQ)^{1/2}]}.
\label{eq:f_dist}
\end{multline}
We use this distance because it measures optimal transport between distributions and is more perceptually linear. We use three semantic clusters and set the threshold to $7.5$.

The threshold of the squared Fr\'echet distance chosen in this sampling strategy controls diversity in the set of styles. A small threshold will lead to little diversity in the results. On the other hand, a large threshold may cause low ranked styles to get sampled, resulting in artifact-prone stylizations. Considering this tradeoff, we use three nearest semantic clusters and set the threshold value to $7.5$.

\Fref{fig:compare_sampling}(b) shows the stylizations our sampling method produces. In comparison, naively sampling the style ranking without enforcing diversity creates multiple results that are visually similar (\Fref{fig:compare_sampling}(a)). On the other hand, increasing the Fr\'echet distance threshold leads to more diversity, but could result in artifacts in the stylizations because of styles at the low-rank end being selected.

\section{Results and Discussion}
\label{sec:results}

We have implemented our stylization technique as a C++ application where the style transfer is parallelized on the CPU. To improve performance, we pre-compute and store the semantic cluster centers of the large photo collection, the style features, and the per-semantic class style ranking. 
At run time, we first extract the CNN feature for the input photograph. The semantic search, style sampling, and style transfer make use of the pre-computed information. 
They take a total of 150 ms (about 40 ms for the CNN feature extraction and 110 ms for style selection and transfer) to create five stylized results from an input image of $1024\times1024$ resolution on an I7 3.4GHz machine. We use the same set of parameters ($\lambda_r=7.5$, $\lambda_l=0.005$, and $\lambda_c=0.05$) to generate all the results in the paper. Please refer to the accompanying video to see a real-time demo of our technique.

\begin{figure*}[t]
	\setlength{\tabcolsep}{1pt}
	\centering
	\includegraphics{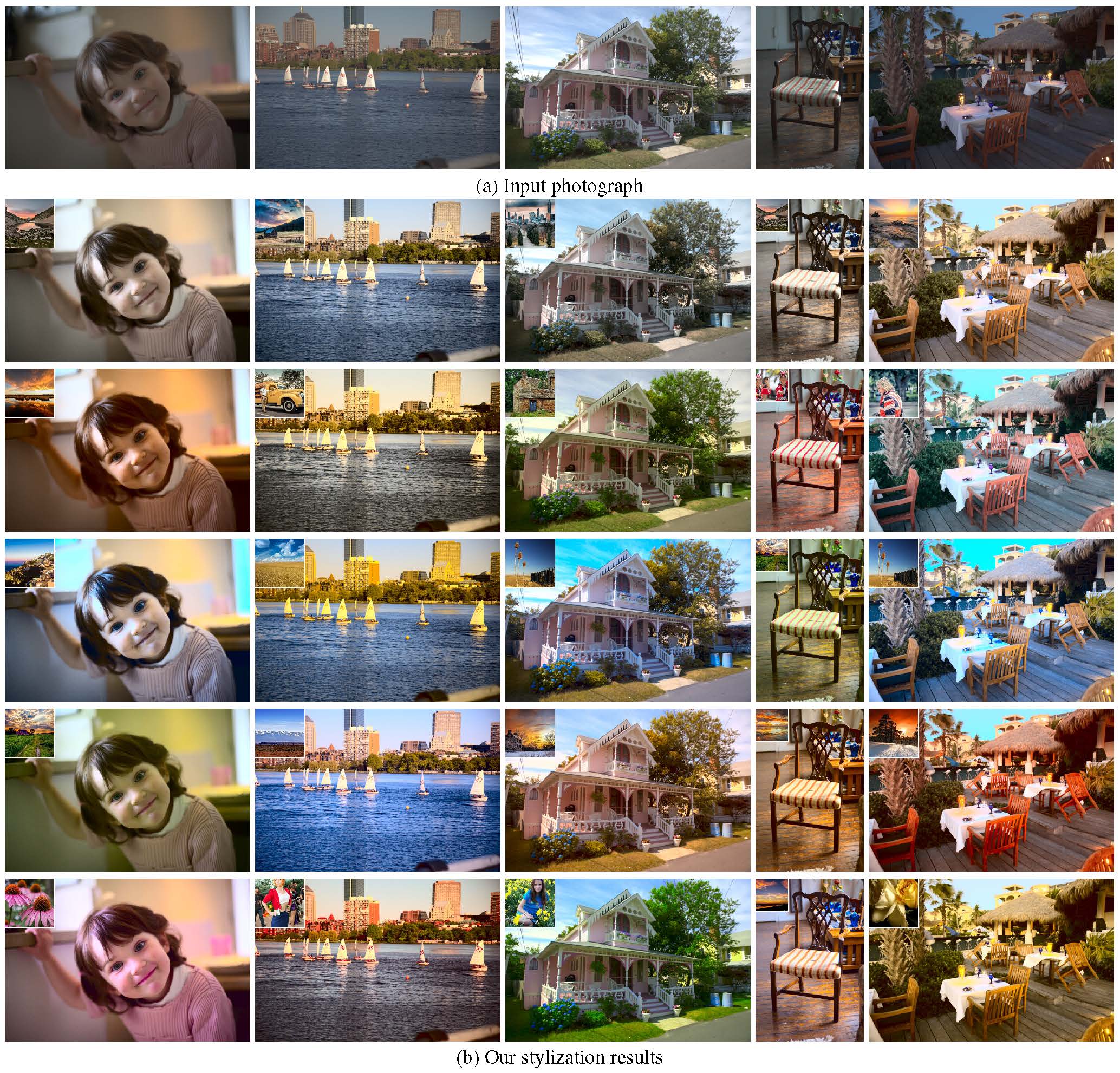}		
	\caption{Our stylization results. The left most images are input photographs and the right images are our automatically stylized results.}
	\label{fig:stylization_result}
\end{figure*}

We have tested our automatic stylization results on a wide range of input images, and show a subset of our results in Figs.~\ref{fig:teaser}~\ref{fig:search_comparisons},~\ref{fig:framework},~\ref{fig:result_pipeline}, and~\ref{fig:stylization_result}. 
Please refer to the supplementary material and video for more examples, comparisons, and a real-time demo of our technique.
As can be seen from these results, our stylization method can robustly capture fairly aggressive visual styles without creating artifacts, and is able to generate diverse stylization results. Figs.~\ref{fig:search_comparisons},~\ref{fig:framework},~\ref{fig:result_pipeline}, and~\ref{fig:stylization_result} also show the automatically chosen style examples that were used to stylize the input photographs. As expected, in most cases, the style examples chosen have different semantics from the input image, but the stylizations are still of high-quality. This verifies the advantage of our method when given only a limited set of stylized exemplars.

\begin{figure*}[t]
	\setlength{\tabcolsep}{1pt}
	\centering 
	\includegraphics[width=0.8\linewidth]{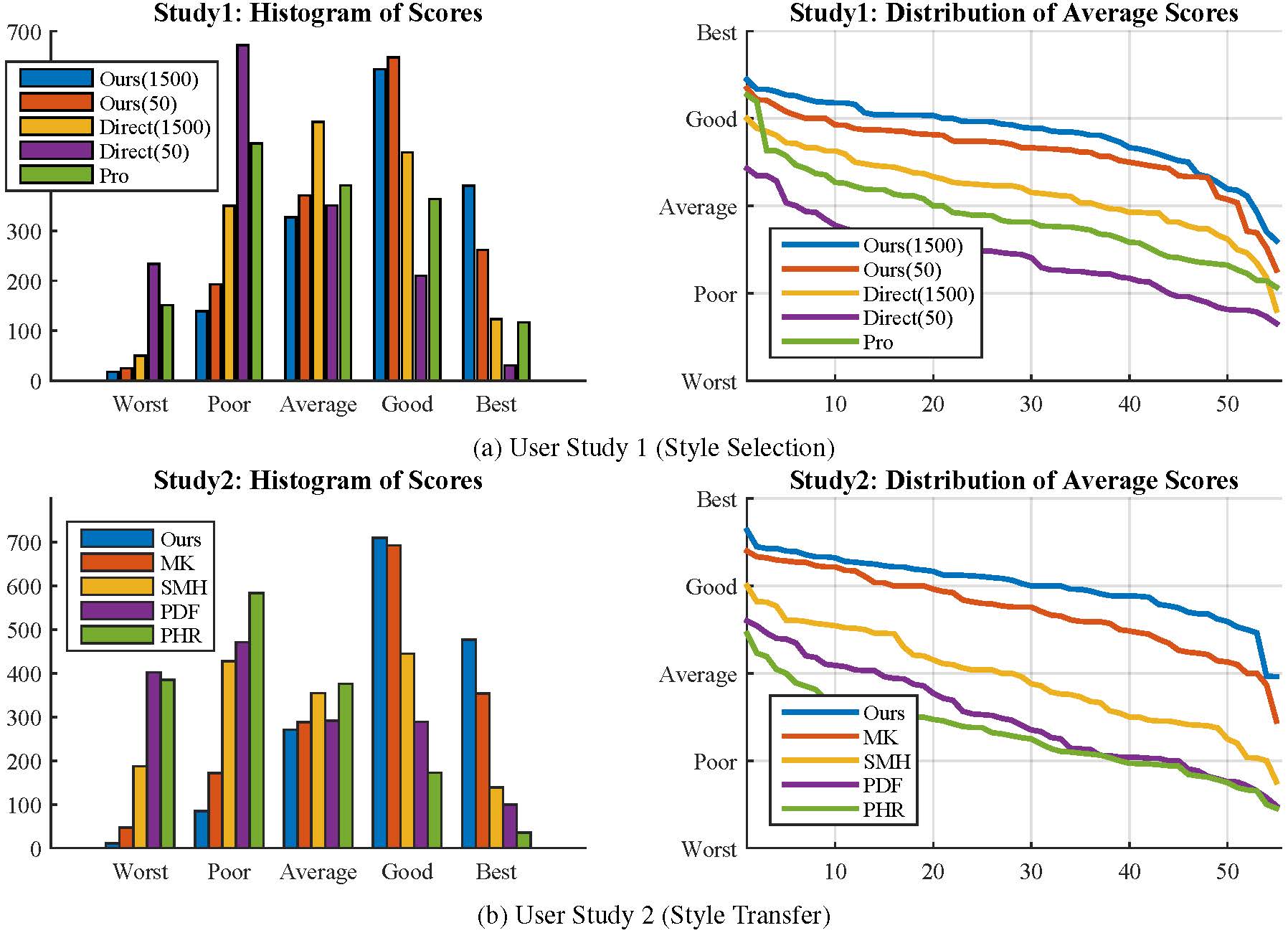}
	\caption{Summary of our two user studies to evaluate our style selection method (a) and our style transfer method (b). For each study, we plot the histogram of user ratings of each tested variant. We also sort the (average) scores achieved by each tested method on each of the $55$ benchmark images of each method and plot these distributions. For both the selection and transfer methods, our algorithms significantly outperform competing methods.}
	\label{fig:user_study_summary}
\end{figure*}

\paragraph{User study} 
Due to the subjective nature of image stylization, we validated our stylization technique through user studies that evaluate our style selection and style transfer strategies. For the study, we created a benchmark dataset of $55$ images -- 50 images were randomly chosen from the FiveK dataset~\cite{bychkovsky2011learning} and the rest were downloaded from Flickr. 
We resized all test images to 500-pixels wide on the long edge and stored them using an 8-bit sRGB JPEG format.

We asked a \textit{professional artist} to create five diverse stylizations for every image in our benchmark dataset as a baseline for evaluation. The artist was told to only use tools that globally edit the color and tone; he used the `Levels', `Curves', `Exposure', `Color Balance', `Hue/Saturation', `Vibrance', and `Black and White' tools in Adobe Photoshop. Creating five different looks for every photograph is challenging even for professional artists. Instead, our artist first constructed $27$ different looks, each of which evoked a particular theme (like `old photo', `sunny', `romantic', etc.), applied all of them to all the images in the dataset, and picked the five diverse styles that he preferred the most.

We performed two user studies. In {\em Study 1}, we evaluated two style selection methods, \textit{our style selection} and \textit{direct semantic search} which directly searches for semantically similar images in the style database. We also explored directly searching in the photo collection using semantic similarity, but its results were consistently poor, which led us to drop this selection method in the larger study. To assess the effect of the size of the style database on the selection algorithm, we tested against two style databases: the full database with 1500 style exemplars, and a small database with 50 style exemplars randomly chosen from the full database. 

We compared five different groups of stylization results including: the reference dataset retouched by a \textit{professional} (henceforth, \textsc{Pro}), our style selection with the full style database (\textsc{Ours 1500}) and the small style database (\textsc{Ours 50}), direct semantic search on the full style database (\textsc{Direct 1500}) and the small style database (\textsc{Direct 50}). For both \textit{our style selection} and \textit{direct semantic search}, we apply the same style sampling in \Sref{sec:style_sampling} to achieve the similar levels of style diversity and create the results using the same style transfer technique (\Sref{sec:style_transfer}). Please see the supplementary material for all these results.

For each image in the benchmark dataset, we showed users five groups of five stylized results (one set each from \textsc{Ours 1500}, \textsc{Ours 50}, \textsc{Direct 1500}, \textsc{Direct 50}, and \textsc{Pro}). Users were asked to rate the stylization quality of each group of results on a five-point Likert scale ranging from 1 (worst) to 5 (best). A total of 37 users participated in this study, and a total of 1498 different image groups were rated, giving us an average of 27.24 ratings per group.

\begin{figure}
	\setlength{\tabcolsep}{1pt}
	\centering 
	\includegraphics{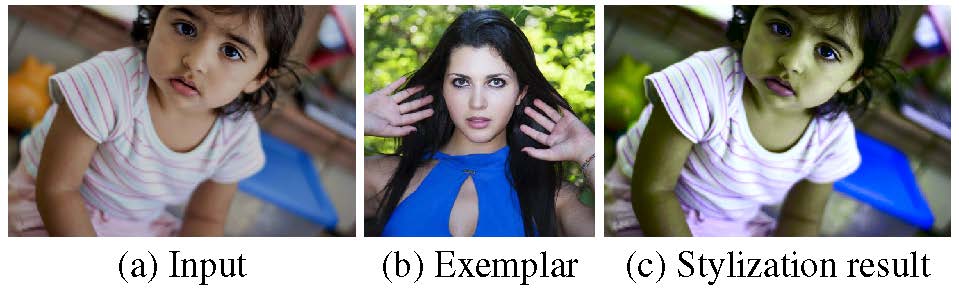}
	\caption{Failure case of direct search.}
	\label{fig:failure_case_direct_search}
\end{figure}

\Fref{fig:user_study_summary}(a) shows the result of Study 1. In this study, \textsc{Ours~1500} ($3.820\pm0.403$) outperforms all the other techniques. We reported the mean of all user ratings and the standard deviation of the average scores of each of the 55 benchmark images. \textsc{Direct~1500} ($3.169\pm0.444$) is substantially worse than \textsc{Ours~1500}. When the style database becomes smaller, the performance of \textit{direct search} drops dramatically ($2.421\pm0.436$ for \textsc{Direct 50}) while \textit{our style selection} stays stable ($3.620\pm0.413$ for \textsc{Ours 50}). We believe that this is a result of our novel two-step style ranking algorithm that is able to learn the mapping between semantic content and style even with very few style examples. On the other hand, direct search fails to find good semantic matches when the size of the style database is reduced significantly. Interestingly, we found that even when direct search finds a semantically meaningful match, this does not guarantee a good style transfer result. An example of this is shown in \Fref{fig:failure_case_direct_search}, where the green in the background of the exemplar image influences the global statistics and causes the girl's skin to take on an undesirable green tone. Our technique aggregates style similarity across many images giving it robustness to such scenarios.

It is also worth noting that \textsc{Pro} ($2.881\pm0.480$) got a lower mean score than \{\textsc{Ours 1500, Ours 50, Direct 1500}\} with the largest standard deviation of scores. We attribute this to two reasons. First, the artist-created filters do not adapt to the content of the image in the same way our example-based style transfer technique does. Second, image stylization tends to be subjective in nature; some of users might be uncomfortable with the aggressive stylizations of a professional, while our style selection is learned from a more `natural' style database and does not have the same level of stylization. 

In {\em Study 2}, we compare our style transfer technique with four different statistics-based style transfer techniques: \textsc{MK}, which computes an affine transform in CIELab~\cite{pitie2007linear}, \textsc{SMH}, which combines three different affine transforms in different luminance bands with a non-linear tone curve~\cite{Bonneel13Video}, \textsc{PDF}, which use 3-d histogram matching in CIELab~\cite{pitie2005n}, and \textsc{PHR}, which progressively reshapes the histograms to make them match~\cite{pouli2011progressive}. We used our implementation for the \textsc{MK} method and used the original authors' code for the other methods.
Style exemplars are chosen by our style selection and these methods are used only for the transfer. We showed users an input photograph, an exemplar, and a randomly arranged set of five stylized images created using the techniques, and asked them to rate the results in terms of style transfer and visual quality on a five-point Likert scale ranging from 1 (worst) to 5 (best). 27 participants from the same pool as (Study 1) participated in this study; they rated 1554 results in total giving us 5.65 ratings per input-style pair and 28.25 rating per input.

\Fref{fig:user_study_summary}(b) shows the result of Study 2. In this study, \textsc{Ours} ($4.002\pm 0.336$) records the best rating, while \textsc{MK} ($3.730\pm0.440$) is ranked second. \textsc{SHM} ($2.949\pm0.545$), \textsc{PDF} ($2.494\pm0.577$), and \textsc{PHR} ($2.286\pm0.452$) are less favored by users. These three techniques have more expressive color transfer models leading to over-fitting and poor results in many cases. This demonstrates the importance of the style transfer technique for high-quality stylization; our technique balances expressiveness and robustness well.

Our evaluation is, to our knowledge, the first extensive evaluation of style transfer techniques. We will release all our benchmark data, including our professionally created dataset, and the results of the different algorithms for other researchers to compare against.

\section{Conclusion}

In this work, we have proposed a completely automatic technique to stylize photographs based on their content. Given a set of target photographic styles, we leverage a large collection of photographs to learn a content-specific style ranking in a completely unsupervised manner. At run-time, we use the learned content-specific style ranking to adaptively stylize images based on their content. Our technique produces a diverse set of compelling, high-quality stylized results. We have extensively evaluated both style selection and transfer components of our technique and studies show that users clearly prefer our results over other variations of our pipeline.

\small
\bibliographystyle{ieee}
\bibliography{ContentAwareStylization}

\end{document}